\documentclass[11pt,a4paper]{article}
\usepackage[T1]{fontenc}
\usepackage[utf8]{inputenc}
\usepackage{amsmath,amssymb}
\usepackage{newtxtext,newtxmath}
\usepackage[margin=2.5cm]{geometry}
\usepackage{setspace}
\usepackage{graphicx}
\usepackage{array,booktabs,adjustbox,multirow}
\usepackage[font=small,labelfont=bf,labelsep=period,justification=justified,singlelinecheck=false]{caption}
\usepackage[section]{placeins}
\usepackage{microtype}
\usepackage{xcolor}
\usepackage{newunicodechar}
\newunicodechar{≤}{\ensuremath{\leq}}
\newunicodechar{≥}{\ensuremath{\geq}}
\newunicodechar{−}{\ensuremath{-}}
\newunicodechar{≈}{\ensuremath{\approx}}
\newunicodechar{±}{\ensuremath{\pm}}
\newunicodechar{→}{\ensuremath{\rightarrow}}
\newunicodechar{←}{\ensuremath{\leftarrow}}
\newunicodechar{↔}{\ensuremath{\leftrightarrow}}
\newunicodechar{∼}{\ensuremath{\sim}}
\newunicodechar{∞}{\ensuremath{\infty}}
\newunicodechar{∈}{\ensuremath{\in}}
\newunicodechar{≠}{\ensuremath{\neq}}
\newunicodechar{⋅}{\ensuremath{\cdot}}
\newunicodechar{∑}{\ensuremath{\sum}}
\newunicodechar{√}{\ensuremath{\surd}}
\newunicodechar{α}{\ensuremath{\alpha}}
\newunicodechar{β}{\ensuremath{\beta}}
\newunicodechar{γ}{\ensuremath{\gamma}}
\newunicodechar{δ}{\ensuremath{\delta}}
\newunicodechar{ε}{\ensuremath{\varepsilon}}
\newunicodechar{κ}{\ensuremath{\kappa}}
\newunicodechar{λ}{\ensuremath{\lambda}}
\newunicodechar{μ}{\ensuremath{\mu}}
\newunicodechar{µ}{\ensuremath{\mu}}
\newunicodechar{π}{\ensuremath{\pi}}
\newunicodechar{σ}{\ensuremath{\sigma}}
\newunicodechar{τ}{\ensuremath{\tau}}
\newunicodechar{ω}{\ensuremath{\omega}}
\newunicodechar{Δ}{\ensuremath{\Delta}}
\newunicodechar{Ω}{\ensuremath{\Omega}}
\newunicodechar{ }{\,}
\newunicodechar{ }{\,}
\usepackage[backend=biber,style=numeric-comp,sorting=none,maxbibnames=6,minbibnames=3,
  giveninits=true,uniquename=false,useprefix=true,doi=true,url=true,isbn=false,eprint=true]{biblatex}
\DeclareNameAlias{default}{family-given}
\DeclareNameAlias{sortname}{family-given}
\renewbibmacro{in:}{\ifentrytype{article}{}{\printtext{\bibstring{in}\intitlepunct}}}
\DeclareFieldFormat[article,inproceedings,incollection,misc,book]{title}{#1\isdot}
\AtEveryBibitem{\clearfield{month}\clearfield{day}\clearfield{urlyear}\clearfield{urlmonth}\clearfield{urlday}\clearlist{language}}

\usepackage{xurl}
\usepackage[colorlinks=true,linkcolor=blue!50!black,citecolor=blue!50!black,urlcolor=blue!50!black,
  bookmarksnumbered=true]{hyperref}
\hypersetup{pdftitle={Bioformer-RetroMAE: retrieval-oriented pretraining of a compact biomedical language model for literature search},pdfauthor={Jialu Li, Li Fang}}
\begin{document}
\begin{flushleft}
\setlength{\parindent}{0pt}
{\LARGE\bfseries\raggedright Bioformer-RetroMAE: retrieval-oriented pretraining of a compact biomedical language model for literature search\par}
\vspace{1.4em}
{\large\raggedright Jialu Li\textsuperscript{1}, Li Fang\textsuperscript{1,*}\par}
\vspace{1em}
{\small\raggedright
\textsuperscript{1}Department of Genetics and Biomedical Informatics, Zhongshan School of Medicine, Sun Yat-sen University, Guangzhou, 510080, China\par\smallskip
\medskip
\textsuperscript{*}Correspondence: \href{mailto:fangli9@mail.sysu.edu.cn}{fangli9@mail.sysu.edu.cn} (L.F.)\par
}
\end{flushleft}

\phantomsection\addcontentsline{toc}{section}{Abstract}
\section*{Abstract}

Dense retrieval is increasingly used for biomedical literature search and retrieval-augmented generation (RAG). Most biomedical bi-encoders are built on \textit{BERT\textsubscript{Base}}-sized encoders pretrained with masked language modeling (MLM). These encoders are costly for corpus-scale indexing and latency-sensitive query encoding, and MLM does not explicitly train the [CLS] vector that a bi-encoder uses to represent a whole text. We present \textit{Bioformer-RetroMAE}, a compact (43M-parameter) retrieval-oriented biomedical encoder obtained by continually pretraining \textit{Bioformer\textsubscript{8L}} with RetroMAE on approximately 32 million PubMed titles and abstracts. After bi-encoder fine-tuning with BM25 hard negatives, we evaluated it on BioASQ Task 12B test questions by retrieving from all PubMed abstracts. \textit{Bioformer-RetroMAE} outperformed \textit{BioBERT} and \textit{PubMedBERT} bi-encoders fine-tuned under identical settings on all metrics, despite having about 40\% of their parameters and producing 512-dimensional rather than 768-dimensional vectors. Compared with \textit{PubMedBERT}, it improved mean average precision (MAP) by 37.0\% and Recall@100 by 29.1\%. Compared with the original \textit{Bioformer\textsubscript{8L}}, RetroMAE pretraining improved MAP by 53.6\% and Recall@100 by 32.8\% without changing model size or inference cost. In hybrid retrieval, fusing \textit{Bioformer-RetroMAE} with BM25 through weighted reciprocal rank fusion improved MAP and NDCG@10 over BM25 alone by 4.9\% and 4.6\%, respectively, larger gains than those obtained by fusing BM25 with \textit{PubMedBERT} or \textit{BioBERT}.

\setcounter{section}{0}
\section{Introduction}\label{sec:1}

PubMed indexes more than 40 million citations and grows by several thousand articles each day. Retrieving relevant evidence from this collection underpins literature curation and evidence synthesis and, increasingly, retrieval-augmented generation (RAG), in which the outputs of large language models (LLMs) are grounded in retrieved documents to reduce hallucinations and make answers verifiable. Unlike offline literature curation, RAG retrieves evidence in real time while the user waits for a response, so retrieval is latency-sensitive. Consequently, the quality, speed and cost of the retrieval step directly affect the reliability and scalability of biomedical question-answering systems and literature-based assistants.

Dense retrieval uses a bi-encoder to map queries and documents into a shared vector space and ranks documents by vector similarity~\cite{Karpukhin2020Dense}. Whereas lexical methods such as BM25~\cite{Robertson2009Probabilistic} rely on exact term overlap, dense retrieval can match semantically related texts that use different wording. Because the two approaches capture complementary relevance signals, they are often combined in hybrid retrieval, for example through reciprocal rank fusion (RRF)~\cite{Cormack2009Reciprocal}. Biomedical dense retrievers, including \textit{MedCPT}~\cite{Jin2023MedCPT}, are typically built on domain-specific encoders such as \textit{PubMedBERT}~\cite{Gu2022Domain}, which adopt the \textit{BERT\textsubscript{Base}} architecture (110M parameters)~\cite{Devlin2019BERT}. For retrieval, encoder size matters in two respects. Offline, every PubMed record (and, increasingly, chunks of PMC full-text articles) must be embedded, and the index must be refreshed as new articles appear. Online, every query must be encoded before search, often repeatedly in multi-hop or iterative RAG workflows, which places query encoding on the critical path of response latency~\cite{Rohanian2024Lightweight,Ngo2025DownscalingFix}. In addition, compact encoders typically have smaller hidden sizes and therefore produce lower-dimensional vectors, which reduce index memory and search cost.

In previous work, we developed \textit{Bioformer}, a family of compact biomedical BERT models pretrained from scratch on PubMed abstracts and PMC full-text articles with a dedicated biomedical vocabulary~\cite{Fang2023Bioformer}. One member of this family, \textit{Bioformer\textsubscript{8L}} (8 layers, hidden size 512, 43M parameters), has approximately 40\% of the parameters of \textit{BERT\textsubscript{Base}} and runs 3 times faster at inference than \textit{BioBERT} and \textit{PubMedBERT}, while achieving comparable accuracy on a range of biomedical NLP tasks~\cite{Fang2023Bioformer}. \textit{Bioformer\textsubscript{8L}} has since been adopted as a backbone in biomedical named entity recognition and normalization tools such as GNorm2~\cite{Wei2023GNorm2}, AIONER~\cite{Luo2023AIONER} and PhenoTagger~\cite{Luo2021PhenoTagger}. These properties make \textit{Bioformer\textsubscript{8L}} an attractive candidate backbone for corpus-scale indexing and low-latency query encoding.

However, \textit{Bioformer\textsubscript{8L}}, like \textit{BioBERT} and \textit{PubMedBERT}, was pretrained with masked language modeling (MLM). MLM predicts masked tokens from token-level representations and does not explicitly train the [CLS] vector that a bi-encoder uses to represent an entire text. Retrieval-oriented pretraining objectives are designed to address this mismatch. RetroMAE~\cite{Xiao2022RetroMAE} pairs the encoder with a lightweight decoder that must reconstruct a heavily masked copy of the input from the encoder's [CLS] vector alone, forcing this vector to capture the content of the full text. Because the decoder is discarded after pretraining, RetroMAE improves single-vector representations without increasing the size or inference cost of the encoder, making it particularly well suited to compact encoders.

Here, we present \textit{Bioformer-RetroMAE}, a compact retrieval-oriented biomedical encoder. We continually pretrained \textit{Bioformer\textsubscript{8L}} with RetroMAE on approximately 32 million PubMed titles and abstracts published through August 2026 and then fine-tuned it as a bi-encoder with BM25 hard negatives. In full-corpus retrieval over all PubMed abstracts on the BioASQ Task 12B golden test sets, we show that (i) \textit{Bioformer-RetroMAE} outperforms \textit{BERT\textsubscript{Base}}-sized biomedical bi-encoders (\textit{BioBERT} and \textit{PubMedBERT}) fine-tuned under identical settings on all evaluation metrics, despite having about 40\% of their parameters; (ii) RetroMAE continual pretraining substantially improves dense retrieval over the original MLM-pretrained \textit{Bioformer\textsubscript{8L}}; and (iii) when BM25 is weighted more heavily in weighted RRF, fusing it with \textit{Bioformer-RetroMAE} yields the strongest pairwise hybrid configuration and improves on BM25 alone in MAP, MAP@10 and NDCG@10 metrics. We release the model weights to support efficient biomedical literature search and RAG pipelines.

\setcounter{section}{1}
\section{Results}\label{sec:2}

\setcounter{section}{2}\setcounter{subsection}{0}
\subsection{Overview of Bioformer-RetroMAE}\label{sec:2.1}

\textit{Bioformer-RetroMAE} was built in two stages (Figure~\ref{fig:1}; see Section~\ref{sec:4}). In the first stage, we continually pretrained the released \textit{Bioformer\textsubscript{8L}} checkpoint with RetroMAE on approximately 32 million PubMed records (title plus abstract) published through August 2026. In RetroMAE, the encoder reads a moderately masked input (masking probability 0.25), and its [CLS] vector is passed to a one-layer decoder that must reconstruct an independently and more aggressively masked version of the same text (masking probability 0.50). The training loss combines the encoder MLM loss and the decoder reconstruction loss with equal weights. The decoder is used only during pretraining, so the resulting encoder keeps the architecture, vocabulary and inference speed of \textit{Bioformer\textsubscript{8L}} (Table~\ref{tab:1}). In the second stage, we fine-tuned separate query and document encoders, both initialized from the RetroMAE-pretrained encoder, on BioASQ Task 12B training questions using an InfoNCE objective with one relevant article and 100 BM25 hard negatives per question. The baseline backbones were fine-tuned under exactly the same setting.

\begin{figure}[!htbp]
\centering
\includegraphics[width=\linewidth,height=0.8\textheight,keepaspectratio]{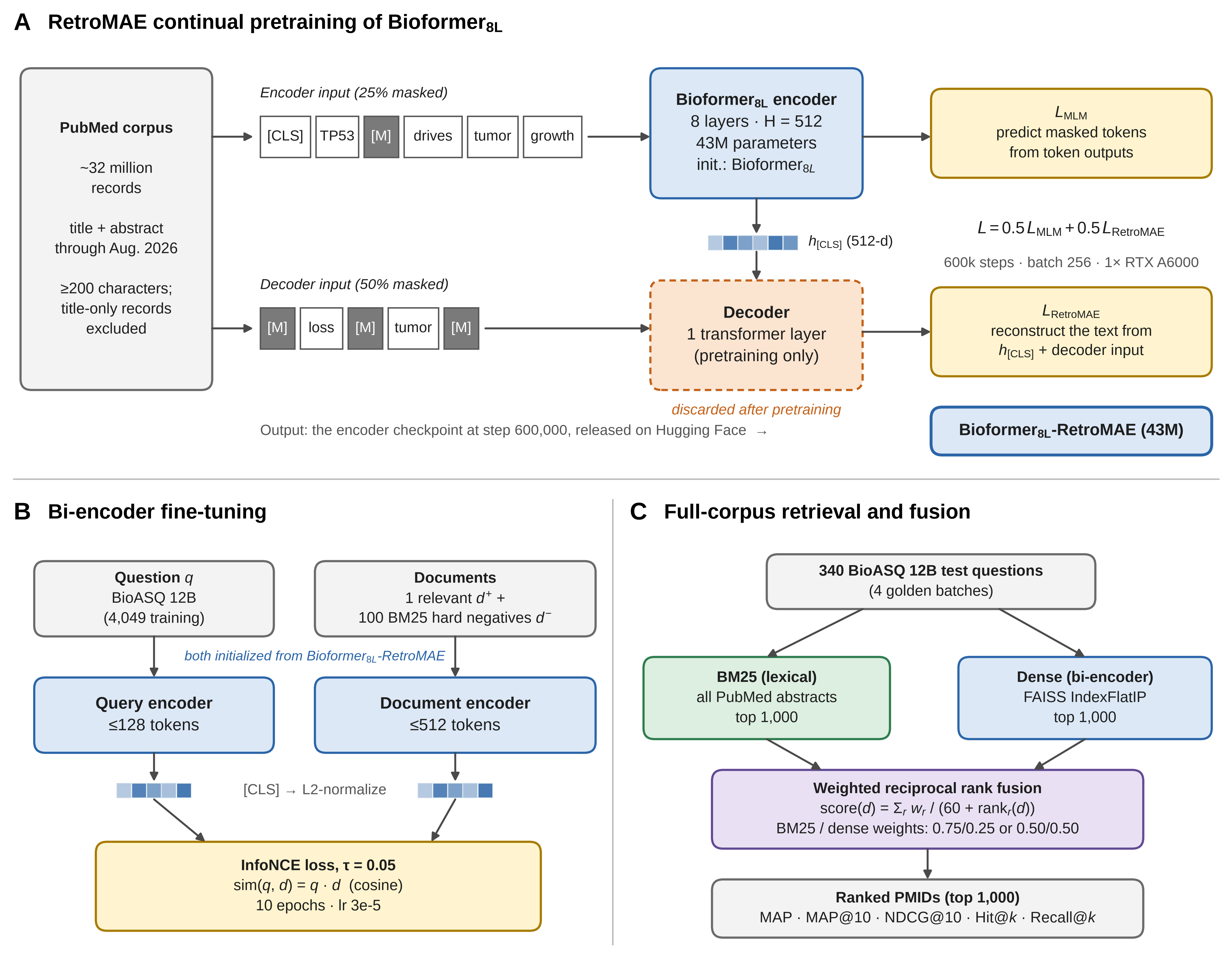}
\setcounter{figure}{0}
\caption{Overview of Bioformer-RetroMAE pretraining, fine-tuning and evaluation. (\textbf{A}) RetroMAE continual pretraining of \textit{Bioformer\textsubscript{8L}} on approximately 32 million PubMed records. The encoder predicts masked tokens from a 25\%-masked input ($\mathcal{L}_{\mathrm{MLM}}$), while a one-layer decoder reconstructs a 50\%-masked copy of the same text from the encoder [CLS] vector ($\mathcal{L}_{\mathrm{RetroMAE}}$). The decoder (dashed outline) is discarded after pretraining, so \textit{Bioformer-RetroMAE} keeps the 43M-parameter size of \textit{Bioformer\textsubscript{8L}}. (\textbf{B}) Bi-encoder fine-tuning. Query and document encoders, both initialized from \textit{Bioformer-RetroMAE}, are trained on BioASQ Task 12B questions with an InfoNCE loss, using one relevant article and 100 BM25 hard negatives per question. Baseline backbones were fine-tuned in the same way. (\textbf{C}) Full-corpus retrieval and fusion. For the 340 BioASQ Task 12B test questions, the top 1,000 PubMed articles retrieved by BM25 and by the dense bi-encoder are merged by weighted reciprocal rank fusion (RRF). MLM, masked language modeling.}\label{fig:1}
\end{figure}

\begin{table}[!htbp]
\centering
\small
\setlength{\tabcolsep}{6pt}
\setcounter{table}{0}
\caption{Backbone encoders compared in this study.}\label{tab:1}
\begin{adjustbox}{max width=\linewidth}
\begin{tabular}{lcccc}
\toprule
 & \textbf{Bioformer-RetroMAE} & \textbf{Bioformer\textsubscript{8L}} & \textbf{BioBERT} & \textbf{PubMedBERT} \\
\midrule
Number of parameters & 43M & 43M & 110M & 110M \\
Number of layers & 8 & 8 & 12 & 12 \\
Hidden size (embedding dimension) & 512 & 512 & 768 & 768 \\
Number of attention heads & 8 & 8 & 12 & 12 \\
Vocabulary size & 32,768 & 32,768 & 28,996 & 30,522 \\
Pretraining objective & MLM + RetroMAE & MLM + NSP & MLM + NSP & MLM \\
\bottomrule
\end{tabular}
\end{adjustbox}
\par\smallskip{\footnotesize\raggedright Notes: MLM, masked language modeling; NSP, next sentence prediction.\par}
\end{table}

RetroMAE continual pretraining converged smoothly over 600,000 optimization steps (approximately 4.8 passes over the corpus). On a held-out set of 10,000 PubMed records, the decoder reconstruction loss decreased from 6.11 at step 10,000 to 3.27 at step 100,000 and 2.97 at step 600,000, with changes of less than 0.01 over the last 100,000 steps. The encoder MLM loss remained stable throughout (1.43 at step 10,000 and 1.41 at step 600,000; masked-token perplexity 4.18 and 4.10, respectively), indicating that the retrieval-oriented objective improved the [CLS] bottleneck representation without degrading the token-level language modeling ability of the encoder.

\setcounter{section}{2}\setcounter{subsection}{1}
\subsection{Dense retrieval performance}\label{sec:2.2}

We trained and evaluated the dense retrievers on BioASQ, a long-running community challenge on large-scale biomedical semantic indexing and question answering~\cite{Nentidis2024OverviewFix}. In its Task B, biomedical experts write English questions that reflect real information needs, and in Phase A, systems must retrieve the PubMed articles relevant to each question. We used Task 12B, the 2024 edition, which provides 5,049 training questions with annotated relevant articles, accumulated over previous editions, and 340 new test questions released in four official golden test batches. Because the questions are written by domain experts and the relevant articles must be found in the entire PubMed collection, Phase A closely reflects the full-corpus retrieval setting of biomedical literature search and RAG pipelines.

We split the 5,049 training questions into 4,049 training and 1,000 validation questions. Each backbone (\textit{Bioformer-RetroMAE}, the original \textit{Bioformer\textsubscript{8L}}, \textit{BioBERT} and \textit{PubMedBERT}) was then fine-tuned as a bi-encoder with separate query and document encoders. For each training question, one relevant article served as the positive, and 100 BM25 hard negatives, highly ranked by BM25 but not annotated as relevant to any training or validation question, served as negatives. Because hard negatives share many terms with the question, they force the model to learn finer relevance distinctions than random negatives would. The model was optimized with an InfoNCE objective to score the positive above all hard negatives. All backbones were fine-tuned with identical training data, hyperparameters and checkpoint-selection criteria, and the checkpoint with the highest mean reciprocal rank on the validation questions was retained (Sections 4.3-4.4). We then evaluated the fine-tuned bi-encoders on the 340 test questions: for each question, each bi-encoder retrieved the top 1,000 articles from an index of all PubMed abstracts (Section~\ref{sec:4.4}). Results are shown in Table~\ref{tab:2}.

\begin{table}[!htbp]
\centering
\footnotesize
\setlength{\tabcolsep}{3pt}
\setcounter{table}{1}
\caption{Dense retrieval performance on the BioASQ Task 12B golden test sets with retrieval from all PubMed abstracts.}\label{tab:2}
\begin{adjustbox}{max width=\linewidth}
\begin{tabular}{lccccccccc}
\toprule
\textbf{Model} & \textbf{MAP} & \textbf{MAP@10} & \textbf{NDCG@10} & \textbf{Hit@1} & \textbf{Hit@10} & \textbf{Hit@100} & \textbf{Recall@1} & \textbf{Recall@10} & \textbf{Recall@100} \\
\midrule
BioBERT & 0.021 & 0.021 & 0.039 & 4.7\% & 16.5\% & 31.8\% & 0.8\% & 2.4\% & 7.1\% \\
PubMedBERT & 0.039 & 0.046 & 0.082 & 13.2\% & 29.7\% & 50.3\% & 1.5\% & 4.7\% & 10.9\% \\
Bioformer\textsubscript{8L} & 0.035 & 0.040 & 0.071 & 10.0\% & 27.1\% & 51.5\% & 1.3\% & 4.2\% & 10.6\% \\
Bioformer-RetroMAE & \textbf{0.053} & \textbf{0.058} & \textbf{0.100} & \textbf{13.5\%} & \textbf{35.0\%} & \textbf{57.9\%} & \textbf{2.3\%} & \textbf{5.7\%} & \textbf{14.1\%} \\
\bottomrule
\end{tabular}
\end{adjustbox}
\par\smallskip{\footnotesize\raggedright Notes:\par 1) Each fine-tuned bi-encoder retrieved the top 1,000 documents per question from the full corpus.\par 2) Bold indicates the best value among the bi-encoders.\par 3) MAP was computed over the top 1,000 retrieved articles; MAP@10 was computed over the top 10 using the official BioASQ Phase A definition (Section~\ref{sec:4.5}).\par 4) Relative improvements reported in the text were computed from unrounded values.\par}
\end{table}

\textit{Bioformer-RetroMAE} achieved the highest scores on all metrics, with a MAP of 0.053, a MAP@10 of 0.058, an NDCG@10 of 0.100, a Hit@10 of 35.0\% and a Recall@100 of 14.1\%. Relative to \textit{PubMedBERT}, the strongest \textit{BERT\textsubscript{Base}}-sized baseline, it improved MAP by 37.0\%, NDCG@10 by 22.1\% and Recall@100 by 29.1\%. The gains over \textit{BioBERT} were larger: 155.1\% in MAP, 157.3\% in NDCG@10 and 97.3\% in Recall@100. These results were obtained with 512-dimensional document vectors, compared with 768-dimensional vectors for the \textit{BERT\textsubscript{Base}}-sized baselines, which reduces the memory footprint of the full-corpus index by one third.

The comparison with the original \textit{Bioformer\textsubscript{8L}} isolates the effect of RetroMAE continual pretraining, because both models share the same architecture and vocabulary and were fine-tuned identically. Before RetroMAE pretraining, \textit{Bioformer\textsubscript{8L}} was slightly behind \textit{PubMedBERT} in MAP (0.035 versus 0.039) and NDCG@10 (0.071 versus 0.082), and comparable in Recall@100 (10.6\% versus 10.9\%). After RetroMAE pretraining, MAP increased by 53.6\%, NDCG@10 by 41.6\% and Recall@100 by 32.8\%, moving the compact encoder from slightly below to clearly above the \textit{BERT\textsubscript{Base}}-sized baselines.

\setcounter{section}{2}\setcounter{subsection}{2}
\subsection{Lexical and hybrid retrieval performance}\label{sec:2.3}

We next compared the dense retrievers with lexical retrieval. BM25~\cite{Robertson2009Probabilistic} is a classical lexical ranking function that scores a document by the query terms it contains, weighting each term by its frequency in the document and its rarity across the corpus and normalizing for document length. Despite its simplicity, BM25 remains a strong baseline that dense retrievers often fail to surpass outside their training domain, and BioASQ is among the benchmarks on which this advantage is most pronounced~\cite{Thakur2021BEIR}. Consistent with this, BM25 alone achieved a MAP of 0.280, an NDCG@10 of 0.391 and a Recall@100 of 57.9\% on the BioASQ Task 12B test questions (Table~\ref{tab:3}), substantially exceeding every dense retriever used alone (Table~\ref{tab:2}). Because dense retrieval matches texts by semantic similarity rather than exact term overlap, we asked whether it could nonetheless add complementary evidence to BM25. We combined BM25 with each dense retriever through weighted RRF and evaluated two settings of the BM25/dense weights (0.75/0.25 and 0.50/0.50).

\begin{table}[!htbp]
\centering
\footnotesize
\setlength{\tabcolsep}{3pt}
\setcounter{table}{2}
\caption{Lexical and hybrid retrieval performance on the BioASQ Task 12B golden test sets with retrieval from all PubMed abstracts.}\label{tab:3}
\begin{adjustbox}{max width=\linewidth}
\begin{tabular}{>{\raggedright\arraybackslash}p{6.1em}lccccccccc}
\toprule
\textbf{Retrieval setting} & \textbf{Model} & \textbf{MAP} & \textbf{MAP@10} & \textbf{NDCG@10} & \textbf{Hit@1} & \textbf{Hit@10} & \textbf{Hit@100} & \textbf{Recall@1} & \textbf{Recall@10} & \textbf{Recall@100} \\
\midrule
Lexical & BM25 & 0.280 & 0.276 & 0.391 & 43.2\% & 80.6\% & 92.1\% & 6.6\% & 24.8\% & 57.9\% \\
\midrule
\multirow{5}{=}{Hybrid (0.75/0.25)} & BioBERT & 0.288 & 0.284 & 0.398 & 44.4\% & 80.9\% & 91.8\% & \textbf{7.2\%} & 25.3\% & 58.1\% \\
 & PubMedBERT & 0.293 & 0.291 & 0.406 & 46.2\% & 80.6\% & 91.8\% & 6.9\% & 25.4\% & 58.0\% \\
 & Bioformer\textsubscript{8L} & 0.289 & 0.289 & 0.405 & \textbf{46.5\%} & \textbf{81.2\%} & 91.8\% & 6.6\% & 25.3\% & 58.2\% \\
 & Bioformer-RetroMAE & \textbf{0.294} & \textbf{0.295} & \textbf{0.409} & \textbf{46.5\%} & 80.6\% & \textbf{92.1\%} & 7.0\% & \textbf{25.6\%} & \textbf{58.6\%} \\
 & All four dense models & 0.300 & 0.301 & 0.412 & 46.8\% & 80.3\% & 91.8\% & 7.7\% & 25.4\% & 58.2\% \\
\midrule
\multirow{5}{=}{Hybrid (0.50/0.50)} & BioBERT & 0.190 & 0.164 & 0.272 & 26.5\% & 75.6\% & \textbf{90.3\%} & 4.3\% & 20.3\% & 49.5\% \\
 & PubMedBERT & 0.217 & 0.209 & 0.318 & 35.0\% & 75.9\% & 89.7\% & 4.7\% & \textbf{21.9\%} & \textbf{51.4\%} \\
 & Bioformer\textsubscript{8L} & 0.203 & 0.192 & 0.304 & 34.4\% & 76.5\% & 90.0\% & 4.4\% & 21.3\% & 50.5\% \\
 & Bioformer-RetroMAE & \textbf{0.219} & \textbf{0.210} & \textbf{0.324} & \textbf{40.9\%} & \textbf{77.1\%} & 89.1\% & \textbf{4.8\%} & 21.6\% & 51.1\% \\
 & All four dense models & 0.298 & 0.298 & 0.412 & 47.1\% & 80.6\% & 91.8\% & 7.3\% & 25.9\% & 58.5\% \\
\bottomrule
\end{tabular}
\end{adjustbox}
\par\smallskip{\footnotesize\raggedright Notes:\par 1) Models were evaluated using the same 340 questions as in Table~\ref{tab:2}. BM25 and each dense retriever returned the top 1,000 documents per question. Hybrid retrieval combined the BM25 and dense retrieval results (Table~\ref{tab:2}) using weighted reciprocal rank fusion with $k_{\mathrm{RRF}}=60$; the weights in parentheses are the BM25/dense weights, and the model name indicates the bi-encoder fused with BM25. For “All four dense models”, BM25 was fused with all four dense retrievers, and the dense-side weight was shared equally among them.\par 2) Bold indicates the best value among the pairwise BM25 + bi-encoder fusions within each weight setting; BM25 alone and the four-model fusion are shown for reference and were not included in this comparison.\par 3) MAP was computed over the top 1,000 retrieved articles; MAP@10 was computed over the top 10 using the official BioASQ Phase A definition (Section~\ref{sec:4.5}).\par}
\end{table}

With BM25/dense weights of 0.75/0.25, every pairwise hybrid configuration outperformed BM25 alone in MAP, MAP@10 and NDCG@10, confirming that lexical and dense rankings provide complementary signals. Among these configurations, \textit{Bioformer-RetroMAE} achieved the highest MAP (0.294), MAP@10 (0.295) and NDCG@10 (0.409), corresponding to relative gains of 4.9\%, 6.9\% and 4.6\% over BM25, respectively. It also achieved the highest Hit@1 (46.5\%, tied with \textit{Bioformer\textsubscript{8L}}), Hit@100 (92.1\%), Recall@10 (25.6\%) and Recall@100 (58.6\%), while its Hit@10 and Recall@1 differed only marginally from the best values.

With equal weights (0.50/0.50), all pairwise hybrid configurations fell below BM25 alone on every metric, reflecting the much weaker standalone performance of the dense retrievers. In this setting, \textit{Bioformer-RetroMAE} showed the smallest degradation in MAP, MAP@10 and NDCG@10, achieving the highest MAP (0.219), MAP@10 (0.210) and NDCG@10 (0.324) among the pairwise configurations, as well as the highest Hit@1 (40.9\%, versus 26.5–35.0\% for the other bi-encoders). Fusing BM25 with all four dense retrievers, with the dense-side weight shared equally among them, was robust to the choice of weights and produced the best overall MAP (0.300 at 0.75/0.25 and 0.298 at 0.50/0.50), MAP@10 (0.301 and 0.298, respectively) and NDCG@10 (0.412 at both settings).

\setcounter{section}{2}
\section{Discussion}\label{sec:3}

In this study, we developed \textit{Bioformer-RetroMAE}, a compact retrieval-oriented biomedical encoder obtained by RetroMAE continual pretraining of \textit{Bioformer\textsubscript{8L}} on approximately 32 million PubMed records. With 43M parameters, about 40\% of the size of \textit{BERT\textsubscript{Base}}-sized biomedical encoders, it achieved the best dense retrieval performance on all evaluation metrics in full-corpus retrieval on the BioASQ Task 12B golden test sets. Its MAP and Recall@100 exceeded those of \textit{PubMedBERT} by 37.0\% and 29.1\%, and those of the original \textit{Bioformer\textsubscript{8L}} by 53.6\% and 32.8\%. Its advantage carried over to hybrid retrieval: with BM25/dense weights of 0.75/0.25, fusing \textit{Bioformer-RetroMAE} with BM25 produced the strongest pairwise hybrid configuration, improving MAP and NDCG@10 over BM25 alone by 4.9\% and 4.6\%.

RetroMAE continual pretraining substantially improved the dense retrieval performance of \textit{Bioformer\textsubscript{8L}}. Before RetroMAE pretraining, the MLM-pretrained \textit{Bioformer\textsubscript{8L}} performed slightly below \textit{PubMedBERT} as a bi-encoder backbone. After RetroMAE pretraining, it outperformed both \textit{BERT\textsubscript{Base}}-sized baselines. This is consistent with the view that MLM leaves the [CLS] vector under-trained for retrieval, and that an objective which explicitly forces the [CLS] vector to reconstruct the text yields better single-vector representations. Because the RetroMAE decoder is discarded after pretraining, these gains come at no cost in inference speed or index size. The encoder retains the architecture of \textit{Bioformer\textsubscript{8L}}, which was previously shown to be approximately 3 times faster than \textit{BERT\textsubscript{Base}}-sized models at inference~\cite{Fang2023Bioformer}, and its 512-dimensional vectors require one third less index memory than the 768-dimensional vectors of \textit{BERT\textsubscript{Base}}-sized encoders. These properties are well suited to indexing tens of millions of PubMed records, to keeping the index up to date as new articles are published, and to latency-sensitive query encoding in interactive or iterative RAG workflows.

A notable observation is that BM25 alone substantially outperformed every dense retriever on the BioASQ test sets (MAP of 0.280 versus at most 0.053, and NDCG@10 of 0.391 versus at most 0.100; Tables 2 and 3). This agrees with previous studies. In the BEIR benchmark, BM25 achieved an NDCG@10 of 0.465 on BioASQ, exceeding all evaluated single-vector dense retrievers, including DPR (0.127), ANCE (0.306), TAS-B (0.383) and GenQ (0.398)~\cite{Thakur2021BEIR}. Unsupervised domain adaptation narrowed but did not close this gap: even after adaptation with generative pseudo-labeling, the best dense model reached an NDCG@10 of 0.442 on the full BioASQ corpus, still below BM25~\cite{Wang2022GPL}. Similarly, the retriever component of \textit{MedCPT}, although trained on large-scale PubMed search logs, achieved an NDCG@10 of 0.332 on BioASQ, and general-domain GTR retrievers with up to 4.8 billion parameters did not exceed 0.324~\cite{Jin2023MedCPT}. Several characteristics of BioASQ may explain the strength of lexical matching. The questions are written by biomedical experts and often contain specific, low-frequency terms, such as gene, protein, drug and disease names, for which exact term matching is highly discriminative. In addition, retrieval over tens of millions of PubMed abstracts requires dense retrievers to separate relevant articles from a very large number of semantically similar but non-relevant articles, which is challenging when only a few thousand training questions are available. Finally, the BioASQ gold standard itself may favor lexical retrieval. Relevant articles are first selected by biomedical experts from the results of keyword queries that they issue to PubMed through the BioASQ annotation tool; after each challenge round, the experts review the articles returned by the participating systems and enrich the gold standard with relevant articles identified by these systems~\cite{Krithara2023BioASQ,Nentidis2024OverviewFix}. Because the expert queries are keyword-based and many systems participating in Task 12B Phase A used BM25 for retrieval~\cite{Nentidis2024OverviewFix}, relevant articles that share few terms with the question are less likely to be found and annotated. Relevant articles retrieved only by a dense retriever may therefore be counted as non-relevant, which would underestimate dense retrieval performance relative to BM25.

Because BM25 was much stronger than the dense retrievers, the benefit of hybrid retrieval depended on how the two components were weighted, consistent with previous analyses of fusion functions~\cite{Bruch2024Analysis}. With equal weights (0.50/0.50), a document ranked first by the dense retriever alone received the same RRF contribution as the top-ranked document from BM25. Because the dense retrievers had low precision at the top ranks, non-relevant articles were promoted above relevant articles retrieved by BM25, and every pairwise hybrid configuration performed worse than BM25 alone (Table~\ref{tab:3}). \textit{Bioformer-RetroMAE} showed the smallest declines in MAP (−21.7\%), MAP@10 (−24.0\%) and NDCG@10 (−17.1\%), compared with 22.5–31.9\%, 24.3–40.5\% and 18.6–30.5\%, respectively, for the other bi-encoders. When the lexical signal was weighted more heavily (0.75/0.25), every pairwise configuration improved on BM25 in MAP, MAP@10 and NDCG@10, although not consistently in Hit@10 or Hit@100, and \textit{Bioformer-RetroMAE} gave the largest improvements in these three metrics. Fusing BM25 with all four dense retrievers yielded the best overall MAP, MAP@10 and NDCG@10 and was insensitive to the weight setting, possibly because averaging over several dense rankings reduces the influence of errors made by any single model. In practice, these results suggest that a compact dense retriever is most useful as a complement to, rather than a replacement for, lexical retrieval in biomedical literature search, and that fusion weights should favor the stronger component.

This study has several limitations. First, we evaluated retrieval on a single benchmark, BioASQ Task 12B, whose expert-written questions favor lexical matching; performance on other query types, such as natural-language clinical questions or queries generated within RAG pipelines, remains to be assessed. Second, we did not include \textit{MedCPT} as a baseline because it was trained on large-scale PubMed search logs that are not publicly available, which would preclude a controlled comparison under identical training data; notably, \textit{MedCPT} is itself initialized from \textit{PubMedBERT}, which we included as a baseline. Third, the bi-encoders were fine-tuned only with BioASQ questions and BM25 hard negatives. Future work could use a cross-encoder as a teacher to train the bi-encoder through knowledge distillation; because a cross-encoder jointly encodes each query–document pair, it can provide more accurate relevance scores, and these soft labels could help the bi-encoder rank relevant articles higher while keeping the efficiency of dense retrieval. Larger and more diverse training data, such as weakly supervised pairs derived from PubMed titles and abstracts or citation links, may further narrow the gap between dense retrieval and BM25. Despite these limitations, \textit{Bioformer-RetroMAE} provides an efficient, publicly available starting point for building dense and hybrid retrievers for biomedical search and RAG pipelines.

\setcounter{section}{3}
\section{Methods}\label{sec:4}

\setcounter{section}{4}\setcounter{subsection}{0}
\subsection{Base model}\label{sec:4.1}

We used the released \textit{Bioformer\textsubscript{8L}} checkpoint (\url{https://huggingface.co/bioformers/bioformer-8L}) as the starting point. \textit{Bioformer\textsubscript{8L}} is a BERT encoder with 8 layers, a hidden size of 512, 8 attention heads and 43M parameters, using a cased biomedical WordPiece vocabulary of 32,768 tokens. It was pretrained from scratch with MLM and next sentence prediction objectives on all PubMed records available as of February 2021 and one million PMC full-text articles; details are given in~\cite{Fang2023Bioformer}. The architecture and vocabulary were kept unchanged in this study.

\setcounter{section}{4}\setcounter{subsection}{1}
\subsection{RetroMAE continual pretraining}\label{sec:4.2}

To improve the single-vector representations produced by \textit{Bioformer\textsubscript{8L}} for dense retrieval without increasing its model size, we conducted retrieval-oriented continual pretraining using the RetroMAE framework~\cite{Xiao2022RetroMAE}.

\textbf{Pretraining corpus.} We collected PubMed records published through August 2026 and excluded records without an abstract (i.e., title-only records) as well as records whose concatenated title and abstract contained fewer than 200 characters, yielding a corpus of approximately 32 million records. The title and abstract of each record were concatenated and treated as a single training instance. Instances were processed independently with dynamic padding and a maximum sequence length of 512 tokens, so that each [CLS] representation corresponded to one PubMed record.

\textbf{Objective.} The encoder was initialized from the released \textit{Bioformer\textsubscript{8L}} checkpoint. Whole-word masking was applied to the encoder input with a masking probability of 0.25. A single-layer transformer decoder was used only during continual pretraining and was discarded before downstream retrieval. Given the final encoder [CLS] representation and an independently masked version of the input, the decoder reconstructed tokens masked at a probability of 0.50. The training objective was an equally weighted combination (0.5 each) of the encoder MLM loss and the decoder reconstruction loss.

\textbf{Optimization.} The model was trained with AdamW using a learning rate of $5 \times 10^{-5}$, a weight decay of 0.01, cosine learning-rate decay and a warmup ratio of 0.10. The per-device batch size was 64, and gradients were accumulated over four steps, resulting in an effective batch size of 256. Training was performed on a single NVIDIA RTX A6000 GPU using BF16 mixed precision and TF32. Every 10,000 optimization steps, we evaluated the model on a held-out set of 10,000 PubMed records and monitored the total validation loss, encoder MLM loss, masked-token perplexity and decoder reconstruction loss. Based on validation-loss convergence, we selected the encoder checkpoint at step 600,000 (approximately 4.8 epochs) for downstream retrieval experiments; this checkpoint is released as \textit{Bioformer-RetroMAE}.

\setcounter{section}{4}\setcounter{subsection}{2}
\subsection{Bi-encoder fine-tuning with BioASQ hard negatives}\label{sec:4.3}

\textbf{Training data.} We fine-tuned the bi-encoders using a custom partition of 5,049 questions derived from BioASQ Task 12B~\cite{Nentidis2024OverviewFix}. The partition comprised 4,049 training questions and 1,000 validation questions. The training questions were associated with 36,783 relevant PubMed articles for which abstracts were available.

\textbf{Hard negatives.} Hard negatives were mined from an index containing all PubMed abstracts. For each training question, BM25~\cite{Robertson2009Probabilistic} retrieved up to 10,000 articles, and the 200 highest-ranked unique PMIDs were retained as candidate negatives. To reduce false-negative contamination, we excluded the union of 43,109 articles annotated as relevant to any of the 5,049 training and validation questions. We then sampled up to 100 negatives per training question uniformly without replacement from the remaining candidates, using a deterministic question-specific random seed. Articles absent from the document store or having fewer than 50 characters in the concatenated title and abstract were excluded. This process yielded 403,568 question–negative pairs, corresponding to a mean of 99.67 negatives per training question.

\textbf{Model and loss.} The bi-encoder used separate query and document encoders. For \textit{Bioformer-RetroMAE}, both encoders were initialized from the encoder checkpoint obtained after 600,000 steps of RetroMAE pretraining; the baseline backbones were initialized from their corresponding pretrained checkpoints: \textit{BioBERT}~\cite{Lee2020BioBERT} from the cased BioBERT-Base v1.2 checkpoint (\url{https://huggingface.co/dmis-lab/biobert-base-cased-v1.2}), \textit{PubMedBERT}~\cite{Gu2022Domain} from the uncased checkpoint pretrained on PubMed abstracts and PMC full-text articles (\url{https://huggingface.co/microsoft/BiomedNLP-BiomedBERT-base-uncased-abstract-fulltext}), and the original \textit{Bioformer\textsubscript{8L}}~\cite{Fang2023Bioformer} from the released checkpoint described in Section~\ref{sec:4.1}. To ensure a fair comparison, all models were fine-tuned using the same training questions, relevant articles, BM25 hard negatives, bi-encoder architecture, optimization settings, validation candidate sets and checkpoint-selection criterion. The final-layer [CLS] hidden state was used as the representation and $L_2$-normalized, producing query and document vectors whose dimensionality equals the encoder hidden size (512 for the \textit{Bioformer\textsubscript{8L}} models and 768 for the \textit{BERT\textsubscript{Base}}-sized baselines). Similarity was calculated as the inner product between normalized vectors and was therefore equivalent to cosine similarity. Each training instance contained one relevant article and 100 BM25 hard negatives. The model was optimized using the InfoNCE objective~\cite{Oord2018Representation} with a temperature of $\tau=0.05$.

\textbf{Optimization.} Fine-tuning was conducted for 10 epochs using a learning rate of $3 \times 10^{-5}$, a weight decay of 0.01, linear learning-rate decay and a warmup ratio of 0.10. Each forward pass processed one question and its associated candidate documents, and gradients were accumulated over 16 question groups before each parameter update. Query and document inputs were limited to maximum lengths of 128 and 512 tokens, respectively. The negative candidates available to each question were reshuffled at the beginning of every epoch. For questions with fewer than 100 retained negatives, additional negatives were sampled with replacement from the available set.

\setcounter{section}{4}\setcounter{subsection}{3}
\subsection{Validation and test settings}\label{sec:4.4}

We used different candidate collections for validation and testing. Validation was performed periodically on question-specific candidate sets, whereas testing was performed by retrieving from the complete indexed corpus.

\textbf{In-training validation.} The validation partition contained 1,000 questions with 8,888 annotated relevant articles in total; individual questions could have multiple relevant articles. For each validation question, one relevant article was used as the positive candidate and ranked against up to 100 question-specific BM25 hard negatives. These fixed candidate sets enabled efficient validation every 250 optimization steps. We monitored mean reciprocal rank (MRR), calculated as the reciprocal rank of the selected relevant article, and Top-1 accuracy, defined as the proportion of questions for which it ranked first.

\textbf{Full-corpus test evaluation.} The checkpoint with the highest validation MRR was selected for test evaluation. The test set contained 340 questions from the four official BioASQ Task 12B golden test batches, and each question was evaluated against an index of all PubMed abstracts. We considered three retrieval settings; results of dense retrieval are reported in Table~\ref{tab:2}, and those of lexical and hybrid retrieval in Table~\ref{tab:3}:

\begin{enumerate}
\item \textit{Lexical retrieval}: BM25 retrieved the top 1,000 documents for each question from the same index and served as the lexical baseline. Duplicate PMIDs in the BM25 output were removed, retaining the first occurrence.
\item \textit{Dense retrieval}: We encoded all documents into dense vectors (512-dimensional for the \textit{Bioformer\textsubscript{8L}} models; 768-dimensional for the \textit{BERT\textsubscript{Base}}-sized baselines) and stored them in a FAISS~\cite{Johnson2021Billion} IndexFlatIP index. For each question, the query encoder retrieved the 1,000 documents with the highest inner-product scores.
\item \textit{Weighted reciprocal rank fusion (RRF)}: To combine lexical and dense retrieval without directly calibrating their scores, we merged the BM25 and dense ranked lists using weighted RRF~\cite{Cormack2009Reciprocal}. For a question $q$, the fused score of a document $d$ was calculated as
\end{enumerate}

\begin{equation*}
\mathrm{score}(d \mid q) = \sum_{r \in \mathcal{S}} \frac{w_r}{k_{\mathrm{RRF}} + \mathrm{rank}_r(d)},
\end{equation*}

where $\mathcal{S}$ is the set of fused runs, $\mathrm{rank}_r(d)$ is the 1-based rank of $d$ in run $r$ (a document absent from a run receives no contribution from that run), $k_{\mathrm{RRF}}=60$, and $\sum_{r \in \mathcal{S}} w_r = 1$. Documents were re-ranked by fused score, with ties broken by the best single-run rank and then by PMID, and the 1,000 documents with the highest fusion scores were retained. We evaluated two fusion schemes: pairwise fusion of BM25 with each dense retriever, and fusion of BM25 with all four dense retrievers, in which the dense-side weight was shared equally among them (i.e., $w_{\mathrm{BM25}} + 4w_{\mathrm{dense}} = 1$). For both schemes, we report two settings of the BM25/dense weights, 0.75/0.25 and 0.50/0.50.

\setcounter{section}{4}\setcounter{subsection}{4}
\subsection{Evaluation metrics}\label{sec:4.5}

We evaluated retrieval performance against the annotated relevant articles with binary relevance, and all metrics were macro-averaged over the 340 test questions. Hit@$k$, with $k \in \{1,10,100\}$, was defined as the proportion of questions with at least one relevant article in the top $k$ retrieved articles, and Recall@$k$ as the fraction of annotated relevant articles retrieved in the top $k$. Mean average precision (MAP) was calculated as the mean of the average precision (AP) of each question over the full retrieved list:

\begin{equation*}
\mathrm{AP}(q_i) = \frac{1}{|\mathcal{D}_i^+|}\sum_{j=1}^{1000} P_i(j)\,\mathrm{rel}_i(j),
\end{equation*}

where $\mathcal{D}_i^+$ is the set of annotated relevant articles for question $q_i$, $P_i(j)$ is the precision of the top $j$ retrieved articles, and $\mathrm{rel}_i(j)$ is 1 if the article at rank $j$ is relevant and 0 otherwise. We also report MAP@10, the official document-retrieval metric of BioASQ Phase A, in which at most 10 articles are returned per question. Following the official BioASQ definition used since BioASQ8~\cite{Nentidis2024OverviewFix}, AP@10 is normalized by the number of relevant articles capped at 10:

\begin{equation*}
\mathrm{AP}@10(q_i) = \frac{1}{\min(10,\,|\mathcal{D}_i^+|)}\sum_{j=1}^{10} P_i(j)\,\mathrm{rel}_i(j).
\end{equation*}

Because of the different normalization, MAP@10 can exceed MAP. We additionally report the normalized discounted cumulative gain at rank 10 (NDCG@10) with binary gains, defined as $\mathrm{NDCG@10} = \mathrm{DCG@10}/\mathrm{IDCG@10}$, where $\mathrm{DCG@10} = \sum_{j=1}^{10} \mathrm{rel}_i(j)/\log_2(j+1)$ and $\mathrm{IDCG@10}$ is the DCG@10 of an ideal ranking.

\phantomsection\addcontentsline{toc}{section}{Acknowledgment}
\section*{Acknowledgment}

This work was supported by the National Natural Science Foundation of China (32300546) (L.F.), the Guangdong Basic and Applied Basic Research Foundation (2025A1515012992) (L.F.), the Prevention and Control of Emerging and Major Infectious Diseases-National Science and Technology Major Project (2027ZD01999500/2027ZD01999503) (L.F.), and the Major Project of Guangzhou National Laboratory (GZNL2024A01003) (L.F.).

\phantomsection\addcontentsline{toc}{section}{Author Contributions}
\section*{Author Contributions}

L.F. conceived and supervised the study, designed the methodology and experiments, performed the RetroMAE pretraining, and wrote and revised the manuscript. J.L. implemented the fine-tuning pipelines, curated the training and evaluation data, performed benchmarking, analyzed the results, and contributed to writing the original draft. All authors read and approved the final manuscript.

\phantomsection\addcontentsline{toc}{section}{Conflict of Interest}
\section*{Conflict of Interest}

None declared.

\phantomsection\addcontentsline{toc}{section}{Data and Code Availability}
\section*{Data and Code Availability}

The \textit{Bioformer-RetroMAE} model is publicly available on Hugging Face (\url{https://huggingface.co/bioformers/Bioformer-8L-RetroMAE}). The base model \textit{Bioformer\textsubscript{8L}} is available at \url{https://huggingface.co/bioformers/bioformer-8L}. The BioASQ Task 12B data are available from the BioASQ challenge website (\url{http://bioasq.org}) upon registration.

\printbibliography[heading=bibintoc,title={References}]

@inproceedings{Karpukhin2020Dense,
  author        = {Karpukhin, Vladimir and Oguz, Barlas and Min, Sewon and Lewis, Patrick and Wu, Ledell and Edunov, Sergey and Chen, Danqi and Yih, Wen-tau},
  title         = {Dense Passage Retrieval for Open-Domain Question Answering},
  booktitle     = {Proceedings of the 2020 Conference on Empirical Methods in Natural Language Processing (EMNLP)},
  pages         = {6769--6781},
  year          = {2020},
  doi           = {10.18653/v1/2020.emnlp-main.550}
}

@article{Robertson2009Probabilistic,
  author        = {Robertson, Stephen and Zaragoza, Hugo},
  title         = {The Probabilistic Relevance Framework: BM25 and Beyond},
  journal       = {Foundations and Trends® in Information Retrieval},
  volume        = {4},
  number        = {1-2},
  pages         = {1--174},
  year          = {2009},
  doi           = {10.1561/1500000019}
}

@inproceedings{Cormack2009Reciprocal,
  author        = {Cormack, Gordon V. and Clarke, Charles L A and Buettcher, Stefan},
  title         = {Reciprocal rank fusion outperforms condorcet and individual rank learning methods},
  booktitle     = {Proceedings of the 32nd international ACM SIGIR conference on Research and development in information retrieval},
  pages         = {758--759},
  year          = {2009},
  doi           = {10.1145/1571941.1572114}
}

@article{Jin2023MedCPT,
  author        = {Jin, Qiao and Kim, Won and Chen, Qingyu and Comeau, Donald C and Yeganova, Lana and Wilbur, W John and Lu, Zhiyong},
  title         = {MedCPT: Contrastive Pre-trained Transformers with large-scale PubMed search logs for zero-shot biomedical information retrieval},
  journal       = {Bioinformatics},
  volume        = {39},
  number        = {11},
  eid           = {btad651},
  year          = {2023},
  doi           = {10.1093/bioinformatics/btad651}
}

@article{Gu2022Domain,
  author        = {Gu, Yu and Tinn, Robert and Cheng, Hao and Lucas, Michael and Usuyama, Naoto and Liu, Xiaodong and Naumann, Tristan and Gao, Jianfeng and Poon, Hoifung},
  title         = {Domain-Specific Language Model Pretraining for Biomedical Natural Language Processing},
  journal       = {ACM Transactions on Computing for Healthcare},
  volume        = {3},
  number        = {1},
  pages         = {1--23},
  year          = {2022},
  doi           = {10.1145/3458754}
}

@inproceedings{Devlin2019BERT,
  author        = {Devlin, Jacob and Chang, Ming-Wei and Lee, Kenton and Toutanova, Kristina},
  title         = {BERT: Pre-training of Deep Bidirectional Transformers for Language Understanding},
  booktitle     = {Proceedings of the 2019 Conference of the North American Chapter of the Association for Computational Linguistics: Human Language Technologies, Volume 1 (Long and Short Papers)},
  pages         = {4171--4186},
  year          = {2019},
  doi           = {10.18653/v1/N19-1423}
}

@article{Rohanian2024Lightweight,
  author        = {Rohanian, Omid and Nouriborji, Mohammadmahdi and Jauncey, Hannah and Kouchaki, Samaneh and Nooralahzadeh, Farhad and {ISARIC Clinical Characterisation Group} and Clifton, Lei and Merson, Laura and Clifton, David A.},
  title         = {Lightweight transformers for clinical natural language processing},
  journal       = {Natural Language Engineering},
  volume        = {30},
  number        = {5},
  pages         = {887--914},
  year          = {2024},
  doi           = {10.1017/S1351324923000542}
}

@misc{Fang2023Bioformer,
  author        = {Fang, Li and Chen, Qingyu and Wei, Chih-Hsuan and Lu, Zhiyong and Wang, Kai},
  title         = {Bioformer: an efficient transformer language model for biomedical text mining},
  eprint        = {2302.01588},
  archiveprefix = {arXiv},
  year          = {2023}
}

@article{Wei2023GNorm2,
  author        = {Wei, Chih-Hsuan and Luo, Ling and Islamaj, Rezarta and Lai, Po-Ting and Lu, Zhiyong},
  title         = {GNorm2: an improved gene name recognition and normalization system},
  journal       = {Bioinformatics},
  volume        = {39},
  number        = {10},
  eid           = {btad599},
  year          = {2023},
  doi           = {10.1093/bioinformatics/btad599}
}

@article{Luo2023AIONER,
  author        = {Luo, Ling and Wei, Chih-Hsuan and Lai, Po-Ting and Leaman, Robert and Chen, Qingyu and Lu, Zhiyong},
  title         = {AIONER: all-in-one scheme-based biomedical named entity recognition using deep learning},
  journal       = {Bioinformatics},
  volume        = {39},
  number        = {5},
  eid           = {btad310},
  year          = {2023},
  doi           = {10.1093/bioinformatics/btad310}
}

@article{Luo2021PhenoTagger,
  author        = {Luo, Ling and Yan, Shankai and Lai, Po-Ting and Veltri, Daniel and Oler, Andrew and Xirasagar, Sandhya and Ghosh, Rajarshi and Similuk, Morgan and Robinson, Peter N and Lu, Zhiyong},
  title         = {PhenoTagger: a hybrid method for phenotype concept recognition using human phenotype ontology},
  journal       = {Bioinformatics},
  volume        = {37},
  number        = {13},
  pages         = {1884--1890},
  year          = {2021},
  doi           = {10.1093/bioinformatics/btab019}
}

@inproceedings{Xiao2022RetroMAE,
  author        = {Xiao, Shitao and Liu, Zheng and Shao, Yingxia and Cao, Zhao},
  title         = {RetroMAE: Pre-Training Retrieval-oriented Language Models Via Masked Auto-Encoder},
  booktitle     = {Proceedings of the 2022 Conference on Empirical Methods in Natural Language Processing},
  pages         = {538--548},
  year          = {2022},
  doi           = {10.18653/v1/2022.emnlp-main.35}
}

@misc{Thakur2021BEIR,
  author        = {Thakur, Nandan and Reimers, Nils and Rücklé, Andreas and Srivastava, Abhishek and Gurevych, Iryna},
  title         = {BEIR: A Heterogenous Benchmark for Zero-shot Evaluation of Information Retrieval Models},
  eprint        = {2104.08663},
  archiveprefix = {arXiv},
  year          = {2021}
}

@inproceedings{Wang2022GPL,
  author        = {Wang, Kexin and Thakur, Nandan and Reimers, Nils and Gurevych, Iryna},
  title         = {GPL: Generative Pseudo Labeling for Unsupervised Domain Adaptation of Dense Retrieval},
  booktitle     = {Proceedings of the 2022 Conference of the North American Chapter of the Association for Computational Linguistics: Human Language Technologies},
  pages         = {2345--2360},
  year          = {2022},
  doi           = {10.18653/v1/2022.naacl-main.168}
}

@article{Krithara2023BioASQ,
  author        = {Krithara, Anastasia and Nentidis, Anastasios and Bougiatiotis, Konstantinos and Paliouras, Georgios},
  title         = {BioASQ-QA: A manually curated corpus for Biomedical Question Answering},
  journal       = {Scientific Data},
  volume        = {10},
  number        = {1},
  eid           = {170},
  year          = {2023},
  doi           = {10.1038/s41597-023-02068-4}
}

@article{Bruch2024Analysis,
  author        = {Bruch, Sebastian and Gai, Siyu and Ingber, Amir},
  title         = {An Analysis of Fusion Functions for Hybrid Retrieval},
  journal       = {ACM Transactions on Information Systems},
  volume        = {42},
  number        = {1},
  pages         = {1--35},
  year          = {2024},
  doi           = {10.1145/3596512}
}

@article{Lee2020BioBERT,
  author        = {Lee, Jinhyuk and Yoon, Wonjin and Kim, Sungdong and Kim, Donghyeon and Kim, Sunkyu and So, Chan Ho and Kang, Jaewoo},
  title         = {BioBERT: a pre-trained biomedical language representation model for biomedical text mining},
  journal       = {Bioinformatics},
  volume        = {36},
  number        = {4},
  pages         = {1234--1240},
  year          = {2020},
  doi           = {10.1093/bioinformatics/btz682}
}

@misc{Oord2018Representation,
  author        = {van den Oord, Aaron and Li, Yazhe and Vinyals, Oriol},
  title         = {Representation Learning with Contrastive Predictive Coding},
  eprint        = {1807.03748},
  archiveprefix = {arXiv},
  year          = {2018}
}

@article{Johnson2021Billion,
  author        = {Johnson, Jeff and Douze, Matthijs and Jegou, Herve},
  title         = {Billion-Scale Similarity Search with GPUs},
  journal       = {IEEE Transactions on Big Data},
  volume        = {7},
  number        = {3},
  pages         = {535--547},
  year          = {2021},
  doi           = {10.1109/TBDATA.2019.2921572}
}

@incollection{Nentidis2024OverviewFix,
  author    = {Nentidis, Anastasios and Katsimpras, Georgios and Krithara, Anastasia and Lima-L{\'o}pez, Salvador and Farr{\'e}-Maduell, Eul{\`a}lia and Krallinger, Martin and Loukachevitch, Natalia and Davydova, Vera and Tutubalina, Elena and Paliouras, Georgios},
  title     = {Overview of {BioASQ} 2024: The Twelfth {BioASQ} Challenge on Large-Scale Biomedical Semantic Indexing and Question Answering},
  booktitle = {Experimental IR Meets Multilinguality, Multimodality, and Interaction. CLEF 2024, Proceedings, Part II},
  editor    = {Goeuriot, Lorraine and Mulhem, Philippe and others},
  series    = {Lecture Notes in Computer Science},
  publisher = {Springer},
  address   = {Cham},
  pages     = {3--27},
  year      = {2024},
  doi       = {10.1007/978-3-031-71908-0_1}
}

@article{Ngo2025DownscalingFix,
  author    = {Ngo, Hoa},
  title     = {Downscaling {BERT} for Efficient Semantic Search of Clinical Terminology},
  journal   = {Studies in Health Technology and Informatics},
  volume    = {333},
  pages     = {52--57},
  year      = {2025},
  doi       = {10.3233/SHTI251575}
}

\end{document}